\documentclass[11pt,a4paper]{article}
\usepackage[hyperref]{emnlp2018}
\usepackage{times}
\usepackage{latexsym}
\usepackage{amsmath}
\usepackage{amsfonts}
\usepackage{xcolor}

\usepackage{url}
\usepackage{booktabs}
\usepackage{graphicx}
\usepackage{enumitem}

\aclfinalcopy % Uncomment this line for the final submission
\title{Privacy-preserving Neural Representations of Text}

\author{Maximin Coavoux \qquad Shashi Narayan \qquad Shay B. Cohen \\ 
  Institute for Language, Cognition and Computation \\
  School of Informatics, University of Edinburgh \smallskip \\
  \texttt{\{mcoavoux,scohen\}@inf.ed.ac.uk,shashi.narayan@ed.ac.uk}
}

\date{}

\begin{document}
\maketitle

\begin{abstract}
  This article deals with adversarial attacks
  towards
  deep learning systems for Natural Language Processing (NLP),
  in the context of privacy protection.
  We study a specific type of attack:
  an attacker eavesdrops on the hidden representations
  of a neural text classifier and tries to recover
  information about the input text.
  Such scenario may arise in situations when
  the computation of a neural network is
  shared across multiple devices, e.g.\ some hidden
  representation is computed by a user's device
  and sent to a cloud-based model.
  We measure the privacy of a hidden representation
  by the ability of an attacker to predict accurately
  specific private information from it
  and characterize the tradeoff between the privacy and the utility
  of neural representations.
  Finally, we propose several defense methods based on modified
  training objectives and show that they improve the privacy
  of neural representations.
\end{abstract}

\section{Introduction}

This article presents an adversarial scenario meant at characterizing
the privacy of neural representations for NLP tasks,
as well as defense methods designed to improve the privacy of those representations.
A deep neural network constructs intermediate hidden representations
to extract features from its input.
Such representations are trained to predict a label,
and therefore should contain useful features for the final prediction.
However, they might also encode information about the input
that a user wants to keep private (e.g.\ personal data)
and can be exploited for adversarial usages.

\begin{figure}
    \resizebox{\columnwidth}{!}{
    \includegraphics{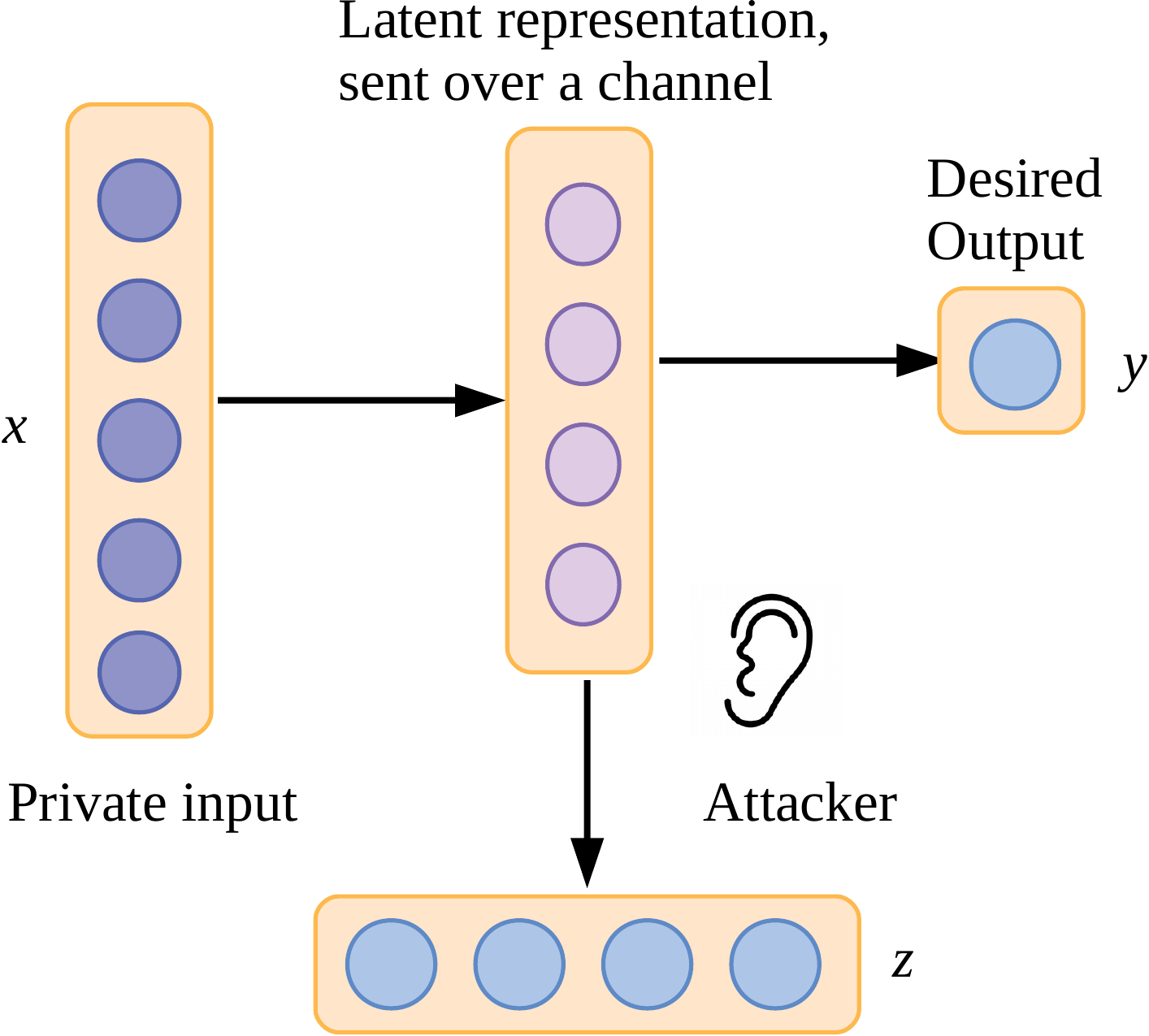}
    }
    \caption{General setting illustration. The main classifier
    predicts a label $y$ from a text $x$, the attacker
    tries to recover some private information $\mathbf z$ contained in $x$
    from the latent representation used by the main classifier.}
    \label{fig:setting}
\end{figure}

We study a specific type of attack on neural representations:
an attacker eavesdrops on the hidden representations
of novel input examples (that are not in the training set)
and tries to recover information about the content
of the input text (Figure~\ref{fig:setting}).
A typical scenario where such attacks would occur is when
the computation of a deep neural net is shared between
several devices \cite{privnet}.
For example, a user's device computes a representation of
a textual input, and sends it a to cloud-based neural network
to obtain, e.g.\ the topic of the text or its sentiment.
The scenario is illustrated in Figure~\ref{fig:setting}.

Private information can take the form of key phrases
\textit{explicitly} contained in the text.
However, it can also be \textit{implicit}. For example,
demographic information about the author of a text
can be predicted with above chance accuracy
from linguistic cues in the text itself
\cite{rosenthal-mckeown:2011:ACL-HLT2011,preoctiucpietro-lampos-aletras:2015:ACL-IJCNLP}.

Independently of its explicitness, some of this private information correlates
with the output labels, and therefore will
be learned by the network.
In such a case, there is a tradeoff between the
utility of the representation (measured by the accuracy
of the network) and its privacy.
It might be necessary to sacrifice some accuracy in order
to satisfy privacy requirements.

However, this is not the case of all private information,
since some of it is not relevant for the prediction of the text label.
Still, private information might be learned incidentally.
This non-intentional and incidental learning
also raises privacy concerns, since an attacker with
an access to the hidden representations, may exploit them to
recover information about the input.

In this paper we explore the following situation:
(i) a \textit{main classifier} uses a deep network to predict a label from textual data;
(ii) an \textit{attacker} eavesdrops on the hidden layers of the network and
tries to recover information about the input text of unseen examples.
In contrast to previous work about neural networks and privacy \cite{DBLP:journals/corr/PapernotAEGT16,DBLP:journals/corr/abs-1802-08232}
we do not protect the privacy of examples from the training set, but
the privacy of unseen examples provided, e.g., by a user.

An example of a potential application would be a spam
detection service with the following constraints:
the service provider does not access verbatim emails sent to users,
only their vector representations.
Theses vector representations should not be usable to gather
information about the user's contacts or correspondents,
i.e.\ protect the user from profiling.

This paper makes the following contributions:\footnote{The source code
used for the experiments described in this paper is available at
\url{https://github.com/mcoavoux/pnet}.}
\begin{itemize}[noitemsep]
 \item We propose a metric to measure the privacy of the neural
 representation of an input for Natural Language Processing tasks.
 The metric is based on the ability of an attacker to recover
 information about the input from the latent representation only.
 \item We present defense methods designed against this type
 of attack. The methods are based on modified training
 objectives and lead to an improved privacy-accuracy tradeoff.
\end{itemize}

\section{Adversarial Scenario}
\label{sec:models}

In the scenario we propose,
each example consists of a triple $(x, y, \mathbf z)$,
where $x$ is a natural language text, $y$ is a single label
(e.g.\ topic or sentiment), and $\mathbf z$ is
a vector of private information contained
in $x$.
Our base setting has two entities: (i) a \textit{main classifier}
whose role is to learn to predict $y$ from $x$,
(ii) an \textit{attacker} who learns to predict $\mathbf z$
from the latent representation of $x$ used by the main classifier.
We illustrate this setting in Figure~\ref{fig:setting}.

In order to evaluate the utility and privacy of a specific model, we proceed in three phases:
\begin{enumerate}[label=Phase \arabic*., wide=10pt,font=\itshape,noitemsep]
    \item Training of the main classifier on $(x, y)$ pairs and evaluation of its accuracy;
    \item Generation of a dataset of pairs $(\mathbf r(x), \mathbf z)$ for the attacker,
    $\mathbf r$ is the representation function of the main classifier ($\mathbf r$ is defined in Section \ref{sec:main});
    \item Training of the attacker's network and evaluation of its performance for measuring privacy.
\end{enumerate}
In the remainder of this section, we describe the main classifier (Section~\ref{sec:main}),
and the attacker's model (Section~\ref{sec:adversary}).

\subsection{Text Classifier}
\label{sec:main}

As our base model, we chose a standard LSTM architecture \cite{Hochreiter:1997:LSM:1246443.1246450}
for sequence classification.
LSTM-based architectures have been applied to many NLP tasks, including
sentiment classification \cite{D16-1058} and text classification \cite{C16-1329}.

First, an LSTM encoder computes a fixed-size representation $\mathbf r(x)$
from a sequence of tokens $x=(x_1, x_2, \dots, x_n)$ projected to an embedding space.
We use $\boldsymbol \theta_{r}$ to denote the parameters used to construct $\mathbf r$.
They include the parameters of the LSTM, as well as the word embeddings.
Then, the encoder output $\mathbf r(x)$ is fed as input
to a feedforward network with parameters $\boldsymbol \theta_{p}$
that predicts the label $y$ of the text, with a softmax output activation.
In the standard setting, the model is trained to minimize the negative log-likelihood of $y$ labels:
\begin{align*}
  \mathcal{L}_m(\boldsymbol \theta_{r}, \boldsymbol \theta_{p}) = \sum_{i=1}^N - \log P(y^{(i)} | x^{(i)}; \boldsymbol \theta_{r}, \boldsymbol \theta_{p}),
\end{align*}
where $N$ is the number of training examples.

\subsection{Attacker's Classifier}
\label{sec:adversary}

Once the main model has been trained, we assume that its parameters
$\boldsymbol \theta_{r}$ and $\boldsymbol \theta_{p}$ are fixed.
We generate a new dataset made of pairs $(\mathbf r(x), \mathbf z(x))$,
where $\mathbf r(x)$ is the hidden representation used by the main model
and $\mathbf z(x)$ is a vector of private categorical variables.
In practice, $\mathbf z$ is a vector of binary variables,
(representing e.g.\ demographic information about the author).
In our experiments, we use the same training examples $x$
for the main classifier and the classifier of the attacker.
However, since the attacker has access to the representation
function $\mathbf r$ parameterized by $\boldsymbol \theta_{r}$,
they can generate a dataset from any corpus
containing the private
variables they want to recover.
In other words, it is not necessary that they have
access to the original training corpus to train their classifier.

The attacker trains a second feedforward network on the new dataset
$\{(\mathbf r(x^{(i)}), \mathbf z^{(i)})\}_{i \leq N}$.
This classifier uses a sigmoid output activation
to compute the probabilities of each binary variable in $\mathbf z$:
\[ P(\mathbf z |\mathbf r(x); \boldsymbol \theta_a) = \sigma(\text{FeedForward}(\mathbf{r}(x))). \]
It is trained to minimize the negative log-likelihood
of $\mathbf z$:
\begin{align*}
 \mathcal{L}_a(\boldsymbol \theta_a) &= \sum_{i=1}^N - \log P(\mathbf z^{(i)} |\mathbf r(x^{(i)}); \boldsymbol \theta_a) \\
  &= \sum_{i=1}^N \sum_{j=1}^K - \log P(\mathbf z^{(i)}_j |\mathbf r(x^{(i)}); \boldsymbol \theta_a),
\end{align*}
assuming that the $K$ variables in $\mathbf z$ are independent.
Since the parameters used to construct $\mathbf r$
are fixed, the attacker only acts upon its
own parameters $\boldsymbol \theta_a$ to optimize
this loss.

We use the performance of the attacker's classifier
as a proxy for privacy.
If its accuracy is high, then an eavesdropper can easily recover
information about the input document.
In contrast, if its accuracy is low (i.e.\ close to
that of a most-frequent label baseline),
then we may reasonably conclude that $\mathbf r$ does
not encode enough information to reconstruct $x$,
and mainly contains information that is useful to
predict $y$.

In general, the performance of a single attacker
does not provide sufficient evidence to conclude
that the input representation $\mathbf r$
is robust to an attack.
It should be robust to any type of reconstruction method.
In the scope of this paper though,
we only experiment with a feedforward network
reconstructor, i.e.\ a powerful learner.

In the following sections, we propose several training
method modifications aimed at obfuscating
private information from the hidden representation
$\mathbf r(x)$.
Intuitively, the aim of these modifications
is to minimize some measure of information
between $\mathbf r$ and $\mathbf z$ to make
the prediction of $\mathbf z$ hard.
An obvious choice for that measure
would be the Mutual Information (MI) between $\mathbf r$ and $\mathbf z$.
However, MI is hard to compute due to the continuous distribution
of $\mathbf r$ and does not lend itself well to stochastic optimization.

\section{Defenses Against Adversarial Attacks}
\label{sec:defenses}

In this section, we present three training methods designed
as defenses against the type of attack we described in Section~\ref{sec:adversary}.
The first two methods are based on two neural networks with rival
objective functions (Section~\ref{sec:adversarial-training}).
The last method is meant at discouraging the model to cluster
together training examples with similar private
variables $\mathbf z$ (Section~\ref{sec:declustering}).

\subsection{Adversarial Training}
\label{sec:adversarial-training}

First, we propose to frame the training of the main classifier
as a two-agent process: the main agent and an adversarial generator,
exploiting a setting similar to Generative Adversarial Networks \cite[GAN,][]{NIPS2014_5423}.
The generator learns to reconstruct examples from the hidden
representation, whereas the main agent learns (i) to perform
its main task (ii) to make the task of the generator difficult.

We experiment with two types of generators:
a classifier that predicts the binary attributes $\mathbf z(x)$
used as a proxy for the reconstruction of $x$ (Section~\ref{sec:multidetasking})
and a character-based language model that directly optimizes
the likelihood of the training examples (Section~\ref{sec:anti-generator}).

\subsubsection{Adversarial Classification: Multidetasking}
\label{sec:multidetasking}

In order not to make $\mathbf r(x)$ a good representation
for reconstructing $\mathbf z$, we make two modifications
to the training setup of the main model (Phase 1):
\begin{itemize}[noitemsep]
 \item We use a duplicate adversarial classifier,
 with parameters $\boldsymbol \theta_a'$,
 that tries to predict~$\mathbf z$ from~$\mathbf r(x)$.
 It is trained simultaneously with the main classifier.
 Its training examples are generated on the fly, and change
 overtime as the main classifier updates its own parameters.
 This classifier simulates an attack during training.
 \item We modify the objective function of the main classifier
 to incorporate a penalty when the adversarial classifier is
 good at reconstructing~$\mathbf z$.
 In other words, the main classifier tries to update
 its parameters so as to confuse the duplicate attacker.
\end{itemize}

Formally, for a single data point $(x, y, \mathbf  z)$,
the adversarial classifier optimizes:
\[ \mathcal{L}_{a'}(x, y, \mathbf z; \boldsymbol \theta_a')_=  - \log P(\mathbf z | \mathbf r(x); \boldsymbol \theta_a'), \]
whereas the main classifier optimizes:
\begin{align*}
\mathcal{L}_m(x, y, \mathbf z; \boldsymbol \theta_r, \boldsymbol \theta_p)_
   =  &- \alpha \log P(y | x;  \boldsymbol \theta_r, \boldsymbol \theta_p)\\
      &- \beta \log P(\neg \mathbf z | \mathbf r(x); \boldsymbol \theta_a').
\end{align*}
The first term of this equation is the log-likelihood of the $y$ labels.
The second term is designed to deceive the adversary.
The hyperparameters $\alpha~>~0$ and $\beta~>0$ control the relative importance
of both terms.

As in a GAN,
the losses of both classifiers are interdependent,
but their parameters are distinct:
the adversary can only update $\boldsymbol \theta_a'$
and the main classifier can only update $\boldsymbol \theta_r$
and $\boldsymbol \theta_p$.

The duplicate adversarial classifier is identical to the classifier
used to evaluate privacy after the main model has been
trained and its parameters are fixed.
However, both classifiers are completely distinct:
the former is used during the training of the main model (Phase~1) to take
privacy into account whereas the latter is used
to evaluate the privacy of the final model (Phase~3),
as is described in Section~\ref{sec:models}.

\subsubsection{Adversarial Generation}
\label{sec:anti-generator}

The second type of generator we use is a character-based LSTM language
model that is trained to reconstruct full training examples.
For a single example $(x; y)$, the hidden state
of the LSTM is initialized with $\mathbf r(x)$, computed by the main model.
The generator optimizes:
\begin{align*}
 \mathcal{L}_g(x, y; \boldsymbol \theta_{\ell}; \boldsymbol \theta_r)
  &=  - \log P(x | \mathbf r(x); \boldsymbol \theta_{\ell}) \\
  &= - \sum_{i=1}^C \log P(x_i | x_{1}^{i-1}, \mathbf r(x); \boldsymbol \theta_{\ell}),
\end{align*}
where $\boldsymbol \theta_{\ell}$ is the set of parameters of the LSTM generator,
$x_i$ is the $i^{th}$ character in the document, and $C$ is the length
of the document in number of characters.
The generator has no control over $\mathbf r(x)$, and optimizes
the objective only by updating its own parameters $\boldsymbol \theta_{\ell}$.

Conversely, the loss of the main model is modified as follows:
\begin{align*}
\mathcal{L}_m(x, y; \boldsymbol \theta_r, \boldsymbol \theta_p)_
   = &- \alpha \log P(y | x;  \boldsymbol \theta_r, \boldsymbol \theta_p) \\
     &- \beta \mathcal{L}_g(x, y; \boldsymbol \theta_{\ell}, \boldsymbol \theta_r).
\end{align*}
The first term maximizes the likelihood of the~$y$ labels whereas
the second term is meant at making the reconstruction difficult by
maximizing the loss of the generator.
As in the loss function described in the previous section, $\alpha$ and $\beta$ control the relative importance of both terms.
Once again, the main classifier can optimize the second term
only by updating $\boldsymbol \theta_r$, since it has no control
over the parameters of the adversarial generator.

A key property of this defense method is that it has no awareness
of what the private variables~$\mathbf z$ are.
Therefore, it has the potential to protect the neural
representation against an attack on any private information.
From a broader perspective, the goal of this defense method
is to specialize the hidden representation $\mathbf r(x)$ to the task
at hand (sentiment or topic prediction) and to avoid learning anything
not relevant to it.

\subsection{Declustering}
\label{sec:declustering}

The last strategy we employ to make the task of the attacker harder
is based on the intuition
that private variables $\mathbf z$ are easier to predict
from~$\mathbf r$ when the main model learns implicitly
to cluster examples with similar $\mathbf z$
in the same regions of the representation space.

In order to avoid such implicit clustering, we add a term
to the training objective of the main model that
penalizes pairs of examples $(x, x')$ that
(i) have similar reconstructions $\mathbf z(x) \approx \mathbf z(x')$
(ii) have hidden representations $\mathbf r(x)$ and $\mathbf r(x')$
in the same region of space. We use the following modified loss
for a single example:
\begin{align*}
  \mathcal{L}_m(x, y, \mathbf z; \boldsymbol \theta_{r}, \boldsymbol \theta_{p})
  = - \log P(y | x; \boldsymbol \theta_{r}, \boldsymbol \theta_{p}) \\
  + \alpha (0.5 - \ell(\mathbf z, \mathbf z')) ||\mathbf r(x) - \mathbf r(x')||_2^2,
\end{align*}
where $(x', \mathbf z')$ is another example sampled uniformly
from the training set,
$\alpha$ is a hyperparameter controlling the importance of the second term,
and $\ell (\cdot, \cdot) \in [0, 1]$ is the normalized Hamming distance.

\section{Experiments}
\label{sec:experiments}

Our experiments are meant to characterize
the privacy-utility tradeoff of neural representations
on text classification tasks,
and evaluating if the proposed defense methods
have a positive impact on it.
We first describe the datasets we used (Section~\ref{sec:datasets}) and
the experimental protocol (Section~\ref{sec:protocol}),
then we discuss the results (Section~\ref{sec:results}).
We found that in the normal training regime, where
no defense is taken into account, the adversary can recover
private information with higher accuracy than a most frequent class baseline.
Furthermore, we found that the defenses we implemented have a positive
effect on the accuracy-privacy tradeoff.

\begin{table}
    \begin{center}
        \begin{tabular}{lrrr}
        \toprule
        Dataset & Train & Dev & Test        \\
        \midrule
        TP US       & 22142 & 2767  & 2767  \\
        TP Germany  & 12596 & 1574  & 1574  \\
        TP Denmark  & 82193 & 10274 & 10274 \\
        TP France   & 9136  & 1141  & 1141  \\
        TP UK       & 48647 & 6080  & 6080  \\
        \midrule
        AG news      & 11657 & 1457 & 1457  \\
        DW corpus    & 5435  & 1772 & 1830  \\
        Blog posts   & 5144  & 642  & 642   \\
        \bottomrule
        \end{tabular}
    \end{center}
    \caption{Sizes of datasets in number of examples.}
    \label{tab:size}
\end{table}

\subsection{Datasets}
\label{sec:datasets}

We experiment with two text classification tasks:
sentiment analysis (Section~\ref{sec:data-sent}) and
topic classification (Section~\ref{sec:data-topic}).
The sizes of each dataset are summarized in Table~\ref{tab:size}.

\subsubsection{Sentiment Analysis}
\label{sec:data-sent}

We use the Trustpilot dataset \cite{Hovy:2015:URS:2736277.2741141}
for sentiment analysis. This corpus contains reviews associated
with a sentiment score on a five point scale, and self-reported information about the users.
We use the five subcorpora corresponding to five areas (Denmark, France, Germany, United Kingdom, United States).

We filter examples containing both the birth year and gender of the author
of the review and use these variables as the private information.
As in previous work on this dataset \cite{hovy:2015:ACL-IJCNLP,P15-2079},
we bin the age of the author
into two categories (`under~35' and `over~45').
Finally, we randomly split each subcorpus into a training set (80\%),
a development set (10\%) and a test (10\%).

\begin{table*}
    \begin{center}
    \begin{tabular}{l cc cc | cc cc}
        \toprule
        & \multicolumn{4}{c|}{Baselines}           & \multicolumn{4}{c}{Best adversaries} \\
        & \multicolumn{2}{c}{Lower bound (most } & \multicolumn{2}{c|}{Upper bound}  &        & & & \\
        & \multicolumn{2}{c}{frequent class)} & \multicolumn{2}{c|}{(trained)}  & \multicolumn{2}{c}{\textsc{+demo}} & \multicolumn{2}{c}{\textsc{raw}} \\
        & Gender & Age & Gender & Age & Gender  & Age & Gender  & Age\\
        \midrule
        TP (Denmark) & 61.6 & 58.4 & 70.5 & 78.0 & 68.5  &  75.3  & 62.0  &  63.4 \\
        TP (France)  & 61.0 & 50.1 & 69.0 & 63.4 & 61.0  &  57.1  & 61.0  &  60.6 \\
        TP (Germany) & 75.2 & 50.9 & 75.2 & 75.2 & 75.2  &  60.4  & 75.2  &  58.6 \\
        TP (UK)      & 58.8 & 56.7 & 70.0 & 76.3 & 66.4  &  63.5  & 59.9  &  61.8 \\
        TP (US)      & 63.5 & 63.7 & 74.1 & 74.8 & 81.3  &  74.9  & 64.7  &  63.9 \\
        Blogs        & 50.0 & 50.3 & 65.7 & 56.1 & -     &  -       &    63.9  &  55.8 \\
        \bottomrule
    \end{tabular}%}
    \end{center}
    \caption{Comparisons between baselines and best adversaries.
    All metrics reported in this table are accuracies.}
    \label{tab:tp-baselines}
\end{table*}

As an additional experimental setting, we use both demographic
variables (gender and age)
as input to the main model. We do so by adding two additional
tokens at the beginning of the input text, one for each variable.
It has been shown that those variables can be used to improve
text classification \cite{hovy:2015:ACL-IJCNLP}.
Also, we would like to evaluate whether the attacker's task is easier
when the variables to predict are \textit{explicitly} in the input,
compared to when these information are only potentially and implicitly in the input.
In other words, this setting simulates the case where private information
may be used by the model to improve classification, but should
not be exposed too obviously.
In the rest of this section, we use \textsc{raw} to denote
the setting where only the raw text is used as input and \textsc{+demo}, the setting
where the demographic variables are also used as input.

\subsubsection{Topic Classification}
\label{sec:data-topic}

We perform topic classification on two genres of documents:
news articles and blog posts.

\paragraph{News article} For topic classification of news article, we use two datasets:
the AG news corpus\footnote{\url{http://www.di.unipi.it/~gulli/AG_corpus_of_news_articles.html}}
\cite{DelCorso:2005:RSN:1060745.1060764}
and the English part of the Deutsche Welle (DW) news corpus \cite{pappas-popescubelis:2017:I17-1}.

For the AG corpus, following \newcite{NIPS20155782},
we construct the dataset by extracting documents belonging to the four most
frequent topics, and use the concatenation of the `title' and `description' fields
as the input to the classifier. We randomly split the corpus into a training set (80\%),
a development set (10\%) and a test set (10\%).
For the DW dataset, we use the `text' field as input, and the standard split.
We kept only documents belonging to the 20 most frequent topics.

The attacker tries
to detect which named entities appear in the input text
(each coefficient in $\mathbf z(x)$ indicates whether a specific named entity occurs in the text).
For both datasets, we used the named entity recognition system from the NLTK package \cite{Bird:2009:NLP:1717171}
to associate each example with the list of named entities that occur in it.
We select the five most frequent named entities with
type `person', and only keep examples containing
at least one of these named entities.
This filtering is necessary to avoid a very unbalanced dataset
(since each selected named entity appears usually in very few articles).

\paragraph{Blog posts} We used the blog authorship corpus
presented by \newcite{841842cf899c42e9b88c7a8cdac64180},
a collection of blog posts associated with the age and gender
of the authors, as provided by the authors themselves.
Since the blog posts have no topic annotation, we ran the LDA algorithm
\cite{Blei:2003:LDA:944919.944937} on the whole collection (with 10 topics).
The LDA outputs a distribution on topics for each blog post.
We selected posts with a single dominating topic ($> 80\%$) and discarded the other posts.
We binned age into two category (under 20 and over 30).
We used the age and gender of the author as the private variables.
These variables have a very unbalanced distribution in the dataset,
we randomly select examples to obtain uniform distributions of private variables.
Finally, we split the corpus into a training set (80\%), a validation set and a test set (10\% each).

\subsection{Protocol}
\label{sec:protocol}

\paragraph{Evaluation}
For the main task, we report a single accuracy measure.
For measuring the privacy of a representation, we compute
the following metrics:
\begin{itemize}[noitemsep]
    \item For demographic variables (sentiment analysis and blog post topic classification):  $1 - X$, where $X$ is the average
    of the accuracy of the attacker on the prediction of gender and age;
    \item For named entities (news topic classification): $1 - F$, where $F$ is an F-score
    computed over the set of binary variables in $\mathbf z$ that indicate the
    presence of named entities in the input example.
\end{itemize}

\paragraph{Training protocol}
We implemented our model using Dynet \cite{dynet}.
The feedforward components (both of the main model and of the attacker)
have a single hidden layer of 64 units with a ReLU activation.
Word embeddings have 32 units.
The LSTM encoder has a single layer of varying sizes, since
it is expected that the amount of information that can be learned
depends on the size of these representations.
We used the Adam optimizer \cite{DBLP:journals/corr/KingmaB14}
with the default learning rate, and 0.2 dropout rate for the LSTM.
We used $\alpha = 0.1$ for the declustering method, based on preliminary
experiments. For the other defense methods, we used $\alpha = \beta = 1$
and did not experiment with other values.

For each dataset, and each LSTM state dimension ($\{8, 16, 32, 64, 128\}$),
we train the main model for 8~epochs (sentiment classification)
or 16~epochs (topic classification),
and select the model with the best accuracy on the development set.
Then, we generate the dataset for the attacker,
train the adversarial model for 16 epochs and select the model with
the worst privacy on the development set (i.e.\ the most successful attacker).

It has to be noted that we select the models that implement defenses
on their accuracy, rather than their privacy or a combination thereof.
In practice, we could also base the selection strategy on a privacy budget:
selecting the most accurate model with privacy above a certain threshold.

\subsection{Results}
\label{sec:results}

This section discusses results for the sentiment analysis task (Section~\ref{sec:results-sentiment})
and the topic classification task (Section~\ref{sec:results-topic}).

\subsubsection{Sentiment Analysis}
\label{sec:results-sentiment}

\paragraph{How private are neural representations?}
Before discussing the effect of proposed defense methods,
we motivate empirically our approach by showing that adversarial models
can recover private information with reasonable accuracy when the attack
is targeted towards a model that implements none of the presented defense methods.

To do so, we compare the accuracy of adversarial models
to two types of baselines:
\begin{itemize}[noitemsep]
    \item As a lower bound, we use the most frequent class baseline.
    \item As an upper bound, we trained a classifier that can
    optimize the hidden representations ($\mathbf r$) for the attacker's tasks.
    In other words, this baseline is trained to predict demographic variables
    from $x$, as if it were the main task.
\end{itemize}

In Table~\ref{tab:tp-baselines}, we compare both baselines to the best
adversary in the two settings (\textsc{raw} and \textsc{+demo}) among the models
trained with no defenses.
First of all, we observe that apart from gender on the German dataset,
the trained baseline outperforms the most frequent class baseline by a wide
margin (8 to 25 absolute difference).
Second of all, the attacker is able to outperform the most frequent class baseline
overall, even in the \textsc{raw} setting.
In more details, for age, the adversary is well over the baseline in all cases
except US. On the other hand, gender seems harder to predict: the adversary outperforms
the most frequent class baseline only in the \textsc{+demo} setting.

The same pattern is visible for the blog post dataset, also presented in the last
line of Table~\ref{tab:tp-baselines}: the best adversaries are 14
points over the baseline for gender and 5 points for age, i.e. almost as
good as a model that can fine tune the hidden representations.

These results justify our approach, since they demonstrate that
hidden representations learn private information about the input,
and can be exploited to recover this information with reasonable accuracy.

\begin{table}[t]
    \resizebox{\columnwidth}{!}{
\begin{tabular}{l|rr|rrrrrrrr}
\toprule
{Corpus} & \multicolumn{2}{c|}{Standard} & \multicolumn{2}{c}{M-Detask.} & \multicolumn{2}{c}{A-Gener.} & \multicolumn{2}{c}{Decl. $\alpha = 0.1$}\\
         & Main & Priv. & Main & Priv. & Main & Priv. & Main & Priv.  \\
\midrule
Germany & 85.1 & 32.2 & -0.6 & -0.3 & -1.3 & \textbf{+0.6} & -0.8 & \textbf{+1.9}\\
baseline & 78.6  & 36.9 \\
\midrule
Denmark & 82.6 & 28.1 & -0.2 & \textbf{+4.4} & -0.1 & \textbf{+6.0} & -0.3 & \textbf{+7.6}\\
baseline & 70.4  & 40.0 \\
\midrule
France & 75.1 & 41.1 & -0.8 & \textbf{+0.7} & -1.4 & -6.4 & -1.5 & -18.2\\
baseline & 69.2  & 44.4 \\
\midrule
UK & 87.0 & 39.3 & -0.5 & \textbf{+0.9} & -0.2 & \textbf{+0.2} & -0.1 & \textbf{+0.3}\\
baseline & 77.1  & 42.2 \\
\midrule
US & 85.0 & 33.9 & -0.1 & \textbf{+2.6} & -0.2 & \textbf{+1.8} & \textbf{+0.7} & \textbf{+2.2}\\
baseline & 79.4  & 36.4 \\
\bottomrule
\end{tabular}
}

    \caption{Results on the test sets of the Trustpilot dataset, \textsc{+demo} setting.
    \textit{Main} is the accuracy on sentiment analysis. \textit{Priv.}\ is the privacy measure
    (i.e.\ the inverse accuracy of the attacker: higher is better, see Section~\ref{sec:protocol}).
    The baselines are most-frequent class classifiers.
    The values reported for the defense methods indicate absolute differences
    with the standard training regime (no defense implemented) for both metrics.}
    \label{tab:results-tp-D}
\end{table}

\begin{table}[t]
    \resizebox{\columnwidth}{!}{
\begin{tabular}{l|rr|rrrrrrrr}
\toprule
{Corpus} & \multicolumn{2}{c|}{Standard} & \multicolumn{2}{c}{M-Detask.} & \multicolumn{2}{c}{A-Gener.} & \multicolumn{2}{c}{Decl. $\alpha = 0.1$}\\
         & Main & Priv. & Main & Priv. & Main & Priv. & Main & Priv.  \\
\midrule
Germany & 85.5 & 32.1 & \textbf{+0.3} & \textbf{+0.5} & -0.8 & \textbf{+0.9} & -1.7 & \textbf{+2.2}\\
baseline & 78.6  & 36.9 \\
\midrule
Denmark & 82.3 & 37.3 & -0.6 & \textbf{+0.6} & -0.1 & -0.3 & -0.2 & -0.1\\
baseline & 70.4  & 40.0 \\
\midrule
France & 72.7 & 40.6 & \textbf{+1.8} & -0.1 & \textbf{+1.9} & -0.4 & -0.3 & -0.1\\
baseline & 69.2  & 44.4 \\
\midrule
UK & 86.9 & 40.1 & -0.2 & \textbf{+1.0} & -0.0 & \textbf{+1.2} & -0.0 & 0.0\\
baseline & 77.1  & 42.2 \\
\midrule
US & 84.5 & 36.1 & -1.1 & \textbf{+0.2} & \textbf{+0.5} & \textbf{+0.1} & \textbf{+0.3} & \textbf{+0.5}\\
baseline & 79.4  & 36.4 \\
\bottomrule
\end{tabular}
}

    \caption{Results on the test sets of the Trustpilot dataset, \textsc{raw} setting.
    See Section~\ref{sec:protocol} and caption of Table~\ref{tab:results-tp-D}
    for details about the metrics.}
    \label{tab:results-tp-noD}
\end{table}

\paragraph{Effect of defenses} We report results for the main task accuracy
and the representation privacy in Table~\ref{tab:results-tp-D}
for the \textsc{+demo} setting and in Table~\ref{tab:results-tp-noD}
for the \textsc{raw} setting.
Recall that the privacy measure (Priv.) is computed by $1-X$ where
$X$ is the average accuracy of the attacker on gender and age predictions.
When this privacy metric is higher, it is more difficult to exploit the
hidden representation of the network to recover information about~$x$.
The `Standard' columns contain the accuracy and privacy of the base
model described in Section~\ref{sec:models}.
The next columns present the absolute variation in accuracy and privacy
for the three defense methods presented in Section~\ref{sec:defenses}:
Multidetasking, Adversarial Generation, and Declustering.
We also report for each corpus the most frequent class baseline
for the main task accuracy, and the privacy of the most frequent class baselines
on private variables (i.e.\ the upper bound for privacy).

The three modified training methods designed as defenses
have a positive effect on privacy.
Despite a model selection based on accuracy, they lead
to an improvement in privacy on all datasets,
except on the France subcorpus.
In most cases, we observe only a small decrease in accuracy, or even
an improvement at times (e.g.\ multidetasking on the Germany dataset, \textsc{raw} setting),
thus improving the tradeoff between the utility and the privacy of the text representations.

\begin{table}
    \resizebox{\columnwidth}{!}{
\begin{tabular}{l|rr|rrrrrrrr}
\toprule
{Corpus} & \multicolumn{2}{c|}{Standard} & \multicolumn{2}{c}{M-Detask.} & \multicolumn{2}{c}{A-Gener.} & \multicolumn{2}{c}{Decl. $\alpha = 0.1$}\\
         & Main & Priv. & Main & Priv. & Main & Priv. & Main & Priv.  \\
\midrule
AG news & 76.5 & 33.7 & -14.5 & \textbf{+14.5} & \textbf{+0.2} & -7.8 & -2.5 & \textbf{+8.6}\\
baseline & 57.8 & & & & & \\
\midrule
DW news & 44.3 & 78.3 & -5.7 & \textbf{+21.7} & \textbf{+5.9} & \textbf{+13.1} & -5.4 & \textbf{+18.4}\\
baseline & 22.1 & & & & & \\
\midrule
Blogs & 58.3 & 40.8 & -0.8 & \textbf{+3.4} & \textbf{+1.1} & \textbf{+0.9} & -0.2 & \textbf{+1.2}\\
baseline & 47.8  & 49.8 \\
\bottomrule
\end{tabular}
}

    \caption{Results for topic classification (test sets).
    See Section~\ref{sec:protocol} and caption of Table~\ref{tab:results-tp-D}
    for details about the metrics.}
    \label{tab:results-ag}
\end{table}

\subsubsection{Topic Classification}
\label{sec:results-topic}

We report results on topic classification in Table~\ref{tab:results-ag}.
\paragraph{News articles}
For the news corpora, the privacy metric is based on the F-score on the binary
variables $\mathbf z$ indicating the presence or absence of a named entity
in the text.
First of all, we observe that defense methods that explicitly
use $\mathbf z$ (i.e.\ multidetasking and declustering), have
a very positive effect on privacy, but also a detrimental effect on the main task.
We hypothesize that this is due to the strong correlations between
the main task labels $y$ and the private information $\mathbf z$.
As a result, improving the privacy of the neural representations
comes at a cost in accuracy.

In contrast, the adversarial generation defense method lead to an improvement
in accuracy, that is quite substantial for the DW corpus.
We speculate that this is due to the secondary term in the objective function
of the main model (Section~\ref{sec:anti-generator})
that helps avoiding overfitting the main task or learning
spurious features.

\paragraph{Blog posts}
On the blog post dataset, the effects are smaller, which we attribute
to the nature of the task of the attacker.
The defense methods consistently improve privacy and, in one case, accuracy.
The best effects on the tradeoff are achieved with the multidetasking and adversarial
generation methods.

\section{Discussion}

The main result of our experiments is that the defenses
we propose improve privacy with usually a small effect, either positive or negative,
on accuracy, thus improving the tradeoff between the utility and the privacy
of neural representations.

An important direction for future work is the choice of a strategy for model selection.
The tradeoff between utility and privacy can be controlled in many ways.
For example, the importance of both terms in the loss functions in
Section~\ref{sec:adversarial-training} can be controlled to favor either
privacy or utility.
In the scope of this paper, we did not perform thorough hyperparameter tuning,
but believe that doing so is important for achieving better results,
since the effects of defense method can be more drastic than desired in some cases,
as exemplified on the news corpora (Table~\ref{tab:results-ag}).

Overall, we found that the multidetasking approach lead to the more stable
improvements and should be preferred in most cases, since it is also
the less computationnally expensive defense.
On the other hand, the adversarial generation method does not require the specification
of private variables, and thus is a more general approach.

\section{Related Work}

The deployment of machine learning in both academic and industrial contexts
raises concerns about adversarial uses of machine learning, as well as concerns
about attacks specifically targeted at these algorithms
that often rely on large amounts of data, including personal data.

More generally, the framework of differential privacy \cite{differential-privacy}
provides privacy guarantees for the problem
of releasing information without compromising confidential data,
and usually involves adding noise in the released information.
It has been applied to the training of deep learning models
\cite{Abadi:2016:DLD:2976749.2978318,DBLP:journals/corr/PapernotAEGT16,2018arXiv180208908P},
and Bayesian topic models \cite{2018arXiv180308471S}.

The notion of privacy is particularly crucial to NLP, since it deals
with textual data, oftentimes user-generated data, that contain a
lot of private information.
For example, textual data contain a lot of signal about authors
\cite{hovy-spruit:2016:P16-2}.
and can be leveraged to predict demographic
variables \cite{rosenthal-mckeown:2011:ACL-HLT2011,preoctiucpietro-lampos-aletras:2015:ACL-IJCNLP}.
Oftentimes, this information is not explicit in the text
but latent and related to the usage of various linguistic traits.
Our work is based on a stronger hypothesis: this latent
information is still present in vectorial representations
of texts, even if the representations have not been supervised
by these latent variables.

\newcite{privnet} study the privacy of unsupervised representations of images,
and measures their privacy with the peak signal to noise ratio
between an original image and its reconstruction by an attacker.
They find a tradeoff between the privacy of the learned representations
and the accuracy of an image classification model that uses these representations as inputs.
Our setting is complementary since it is applied to NLP tasks, but explores
a similar problem in the case of representations learned with a task supervision.

A related problem is the unintended memorization of private data from the training
set and has been addressed by
\newcite{DBLP:journals/corr/abs-1802-08232}.
They tackle this problem
in the context of text generation (machine translation, language modelling).
If an attacker has access to e.g.\ a trained language model, they are likely
to be able to generate sentences from the training set, since the language
model is trained to assign high probabilities to those sentences.
Such memorization is problematic when the training data contains private
information and personal data.
The experimental setting we explore is different from these works:
we assume that the attacker has access to a hidden layer of the network
and tries to recover information about an input example that is not in the
training set.

In a recent study, \newcite{P18-2005} proposed a method
based on GAN designed to improve the robustness and privacy of neural
representations, applied to part-of-speech tagging and sentiment analysis.
They use a training scheme with two agents similar to our multidetasking
strategy (Section~\ref{sec:multidetasking}), and found that it made
neural representations more robust and accurate.
However, they only use a single adversary to alter the training of the main model
and to evaluate the privacy of the representations, with the risk of
overestimating privacy.
In contrast, once the parameters of our main model are fixed,
we train a new classifier from scratch to evaluate privacy.

\section{Conclusion}

We have presented an adversarial scenario and used it to measure the privacy
of hidden representations in the context of two NLP tasks: sentiment analysis
and topic classification of news article and blog posts.
We have shown that in general, it is possible for
an attacker to recover private variables with higher than chance accuracy,
using only hidden representations.
In order to improve the privacy of hidden representations,
we have proposed defense methods based on modifications
of the training objective of the main model.
Empirically, the proposed defenses lead to models with a better privacy.

\section*{Acknowledgments}
We thank the anonymous reviewers and members of the Cohort
for helpful feedback on previous versions of the article.
We gratefully acknowledge the support of the European Union under the Horizon
2020 SUMMA project (grant agreement 688139), and the support of Huawei Technologies.

\bibliography{emnlp2018}
\bibliographystyle{acl_natbib_nourl}

\end{document}